\documentclass[letterpaper,10pt,conference]{ieeeconf}
\IEEEoverridecommandlockouts

\usepackage{amsmath,amssymb,amsthm,mathtools}
\usepackage{booktabs}
\usepackage{array}
\usepackage{tabularx}
\usepackage{graphicx}
\IfFileExists{figures/game_results1.pdf}{%
  \graphicspath{{figures/}}%
}{%
  \graphicspath{{figures/}}%
}
\usepackage{algorithm}
\usepackage{flafter}
\usepackage{placeins}
\usepackage{algpseudocode}
\usepackage{cite}
\PassOptionsToPackage{hyphens}{url}
\usepackage{url}
\usepackage[hidelinks]{hyperref}

\newcommand{\cY}{\mathcal{Y}}
\newcommand{\cR}{\mathcal{R}}
\newcommand{\btheta}{\boldsymbol{\theta}}
\newcommand{\Prob}{\mathbb{P}}
\newcommand{\E}{\mathbb{E}}
\newcommand{\doop}{\mathrm{do}}
\newcommand{\MG}{M_G}
\newcommand{\tablefont}{\footnotesize}
\newcolumntype{C}{>{\centering\arraybackslash}X}
\usepackage[hyphens]{url}

\theoremstyle{definition}
\newtheorem{theorem}{Theorem}
\newtheorem{proposition}[theorem]{Proposition}
\newtheorem{corollary}[theorem]{Corollary}
\newtheorem{assumption}{Assumption}
\newtheorem{definition}{Definition}

\title{\LARGE Dynamics-Induced Commitment in Learning-Based Robotic Penalty Kicks}

\author{Ruize Geng$^{*,1}$, Hao E. Zhang$^{*,1,2}$, Yisen Li$^{2}$, Yikai Wang$^{1}$, H. Eric Tseng$^{\dagger,2}$ and Ding Zhao$^{\dagger,1}$
\thanks{$^{*}$Equal contribution.}
\thanks{$^{\dagger}$Correspondance to Ding Zhao (dingzhao@cmu.edu), H. Eric Tseng (hongtei.tseng@uta.edu) and Hao E. Zhang (haoz4@andrew.cmu.edu).}
\thanks{$^{1}$Ruize Geng, Hao E. Zhang, Yikai Wang and Ding Zhao are with Carnegie Mellon University.
        (email: rgeng3@jh.edu; haoz4@andrew.cmu.edu; yikaiw2@andrew.cmu.edu; dingzhao@cmu.edu)}%
\thanks{$^{2}$Hao E. Zhang, Yisen Li and H. Eric Tseng are with the University of Texas at Arlington.
        (email: yisen03@upenn.edu; hongtei.tseng@uta.edu)}%
}

\begin{document}
\maketitle
\thispagestyle{empty}
\pagestyle{empty}

\begin{abstract}
Learning in robotic games is constrained not only by strategic information but also by what the body can still execute.
We study this coupling in a hierarchical humanoid--quadruped penalty system in which game-level self-play policies command fixed soccer whole-body controllers (S-WBCs). The humanoid shooting skill is initialized from self-collected motion-capture data, whereas the quadruped saving skill is learned by reinforcement learning.
We introduce dynamics-induced commitment mapping (DIC-Map), a body-grounded analysis that estimates continuation capability, identifies the first persistent loss of a terminal alternative, and tests whether the remaining interaction admits a reduced zero-sum game.
For symmetric terminal alternatives, the reduced game yields a closed-form bound on optimal strategy concentration determined by the responder's value of deferring.
We further show that, when the responder acts through an estimator, equal response values eliminate the direct terminal-allocation gradient and leave an estimator-mediated first-order learning channel.
Experiments locate commitment about $0.29$\,s before contact, and changing only ball speed shifts deferral coverage.
Across four responder policies, replacing the estimator raises save rate from $0.240$ to $0.472$, whereas a comparable gain in read accuracy obtained by waiting raises it only to $0.246$.
Posterior analysis is used for the equilibrium comparison because the available coverage terms are observational proxies. Project website: \url{https://chris-ruizegeng.github.io/penaltykick/}
\end{abstract}

\section{Introduction}
\label{sec:intro}

Learning-based robotic competition often separates strategic decisions from whole-body execution.
Recent robot-soccer systems demonstrate learned whole-body skills \cite{haarnoja2024soccer} and team-level coordination \cite{su2025quadruped}.
A strategic policy chooses what to do, while a high-rate whole-body controller determines what the robot can still physically realize.
This separation supports agile contact-rich skills but creates a game-theoretic mismatch.
A high-level policy may continue to assign probability to a nominal alternative after momentum \cite{zhang2025bi}, contact, actuation limits, or delay have made its terminal outcome unreachable.
The effective strategic action set can contract before either agent explicitly commits.
In embodied games, dynamics do not merely constrain execution. They determine which strategic alternatives still exist.

\begin{figure}[t]
    \centering
    \includegraphics[width=0.95\linewidth]{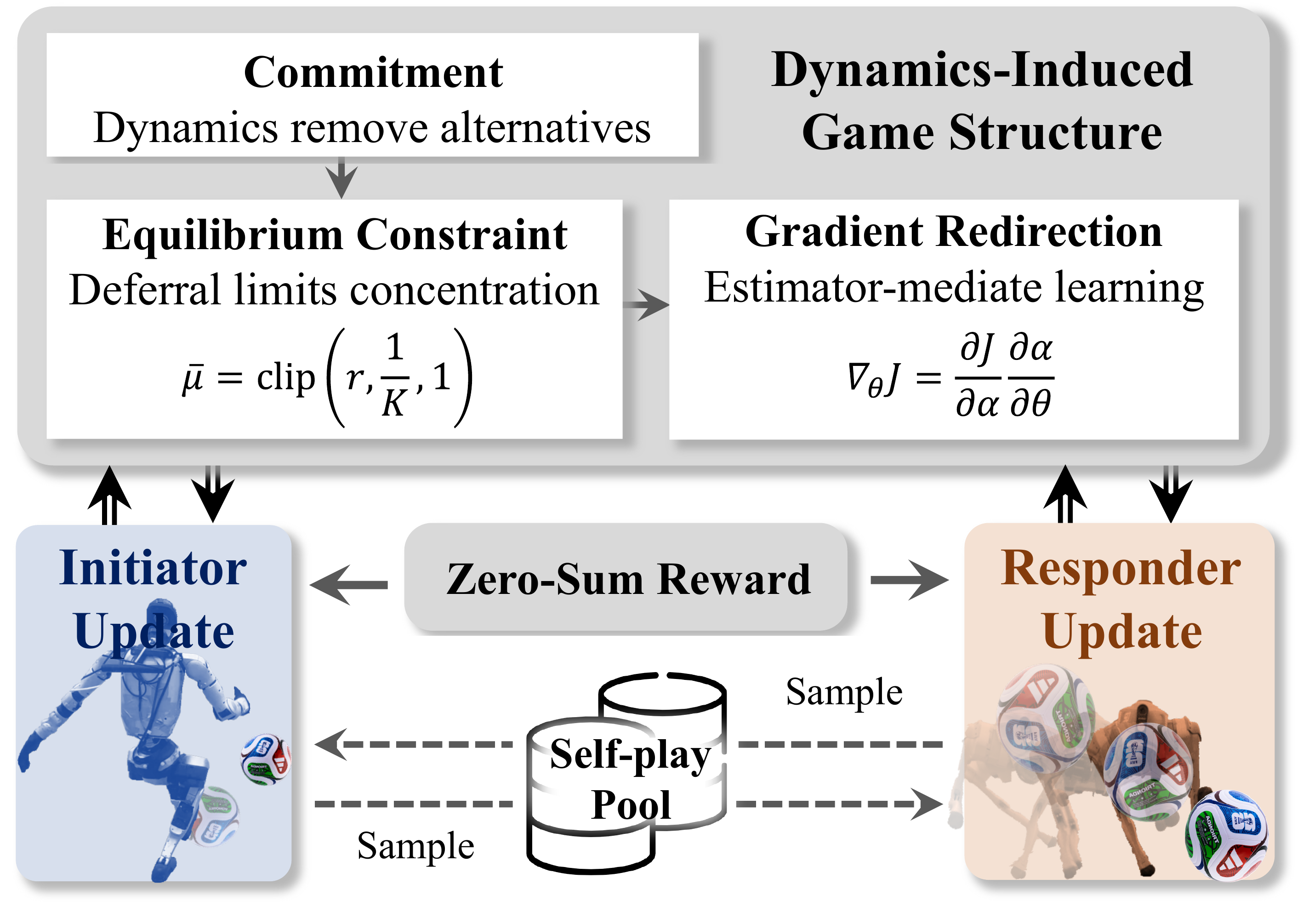}
    \caption{Overview of the proposed game-learning framework, where physical dynamics induce commitment, the responder's deferral option bounds equilibrium concentration, and estimator response values determine the active first-order learning channel during zero-sum self-play.}
    \label{fig:game-overview}
\end{figure}

Game-theoretic models place strategic timing at information or decision events \cite{ghimire2026football}.
Reachability analysis instead asks which outcomes remain dynamically feasible \cite{mitchell2005reachability}, while delay-aware learning shows that latency alters closed-loop feasibility \cite{chen2020delay}.
In embodied competition these mechanisms are coupled. The initiator may lose a terminal alternative while the responder is still deciding whether to commit or defer, and the value of waiting depends on how much physical response authority remains \cite{zhang2026halo}.
The resulting timing constrains the terminal equilibrium and the policy perturbations that can change payoff. We formalize this mechanism as dynamics-induced commitment and operationalize it with dynamics-induced commitment mapping (DIC-Map), see Fig.~\ref{fig:game-overview}.

DIC-Map estimates continuation capability from physical rollouts, defines commitment by the first persistent loss of a terminal alternative, and uses initiator--responder timing to test whether the remaining interaction admits a reduced zero-sum game.
For symmetric alternatives, the responder's value of deferring gives a closed-form upper bound on optimal terminal concentration.
When the responder acts through an outcome estimator, the reduced model also identifies when terminal reallocation is first-order flat and learning is driven through estimator sensitivity.
We instantiate the analysis in a heterogeneous humanoid--quadruped penalty game. Fixed soccer whole-body controllers (S-WBCs) provide executable shooting and saving skills, while game-level actor--critics revise strategic commands through zero-sum self-play over a shared policy pool \cite{lanctot2017psro}.
The humanoid skill is initialized from self-collected motion capture, whereas the quadruped saving skill is learned from scratch.
Because the S-WBCs are frozen during self-play, strategic commands can continue to change even as the body's executable alternatives contract.

We evaluate the mechanisms in simulation with a $29$-degree-of-freedom humanoid initiator and a quadruped responder, with complementary hardware trials.
The final shot side can be read with about $0.92$ accuracy $0.5$\,s before contact, while both terminal alternatives remain physically reachable until about $0.29$\,s before contact.
Changing only ball speed shifts deferral coverage. Replacing the estimator raises save rate from $0.240$ to $0.472$, whereas a comparable gain in read accuracy obtained by waiting raises it only to $0.246$.
Self-play exhibits increased within-episode target revision and more concentrated terminal allocation. Hardware trials show both a late direction change that misdirects the responder and a successful anticipatory save.

The paper makes three contributions: 1) dynamics-induced commitment is formulated as the body-grounded contraction of strategic alternatives, with DIC-Map estimating continuation capability, identifying persistent commitment time, and testing a reduced endgame from whole-body rollouts; 2) deferral value is shown to impose a closed-form ceiling on optimal terminal concentration, while equal response values across terminal outcomes eliminate the direct terminal-allocation gradient and redirect the first-order effect through estimator sensitivity; 3) these mechanisms are evaluated in a heterogeneous humanoid--quadruped penalty game with learned whole-body skills and zero-sum self-play, revealing a readability--commitment gap and estimator-versus-waiting asymmetry, together with late-revision and anticipatory interactions on hardware.

\section{Related Work}

\subsection{Asymmetric Games and Strategic Timing}
Repeated games with incomplete information model belief updates under private information \cite{zhang2025multi}, with dual formulations characterizing related repeated-game structure \cite{demeyer1996repeated}.
Differential-game formulations extend asymmetric information to continuous time \cite{cardaliaguet2007differential}, including zero-sum games \cite{ghimire2024state}.
Ghimire et al.~derive a critical revealing time in a continuous-action football game \cite{ghimire2026football}.
Endogenous-commitment models assign value to delaying a choice \cite{caruana2008endogenous}, and preemption models study the strategic value of acting first \cite{fudenberg1985preemption}.

\subsection{Reachability, Delay, and Physical Feasibility}
Reachability analysis characterizes dynamically feasible sets in continuous dynamic games \cite{mitchell2005reachability}.
Delay-aware learning shows that sensing and actuation latency alter closed-loop feasibility \cite{chen2020delay}.
DIC-Map instead uses continuation capability estimated from branched rollouts over a stated controller class rather than a certified backward-reachable set.
This makes strategic availability measurable in high-dimensional whole-body systems.
We score strategic quantities from realized terminal states rather than command-level proxies, consistent with concerns about proxy misspecification \cite{amodei2016concrete} and specification gaming \cite{krakovna2020specification}.

\subsection{Multi-Agent Learning and Opponent Estimation}
Fictitious self-play trains policies against iterated responses \cite{heinrich2015fsp}.
Policy-space response methods extend this idea to policy populations \cite{zhang2026cognition}, while large-scale self-play has demonstrated complex strategic adaptation \cite{vinyals2019alphastar}.
PPO is a common actor--critic choice \cite{schulman2017ppo}, with MAPPO showing its effectiveness in multi-agent settings \cite{yu2022mappo}.
Opponent-shaping methods reason about another learner's update \cite{foerster2018lola}, with model-free extensions removing the differentiable opponent model \cite{lu2022mfos}.
DIC-Map does not shape the opponent because the responder estimator is frozen during policy learning.
The relevant effect is distribution shift against a fixed predictor, related to performative prediction \cite{perdomo2020performative}, strategic classification \cite{hardt2016strategic}, and dataset aggregation \cite{ross2011dagger}.
Recent embodied soccer systems learn whole-body behavior with reinforcement learning \cite{haarnoja2024soccer,zhang2026interaction}.
Related work extends this setting to egocentric interaction \cite{tirumala2025egocentric} and quadruped team play \cite{su2025quadruped}.
These works establish soccer skills and strategic interaction. Our question is different: when does closed-loop whole-body execution remove a strategic alternative, and how does that change the effective game and learning signal?

\section{Methodology}
\label{sec:framework}

\begin{figure*}[t]
    \centering
    \includegraphics[width=0.95\textwidth]{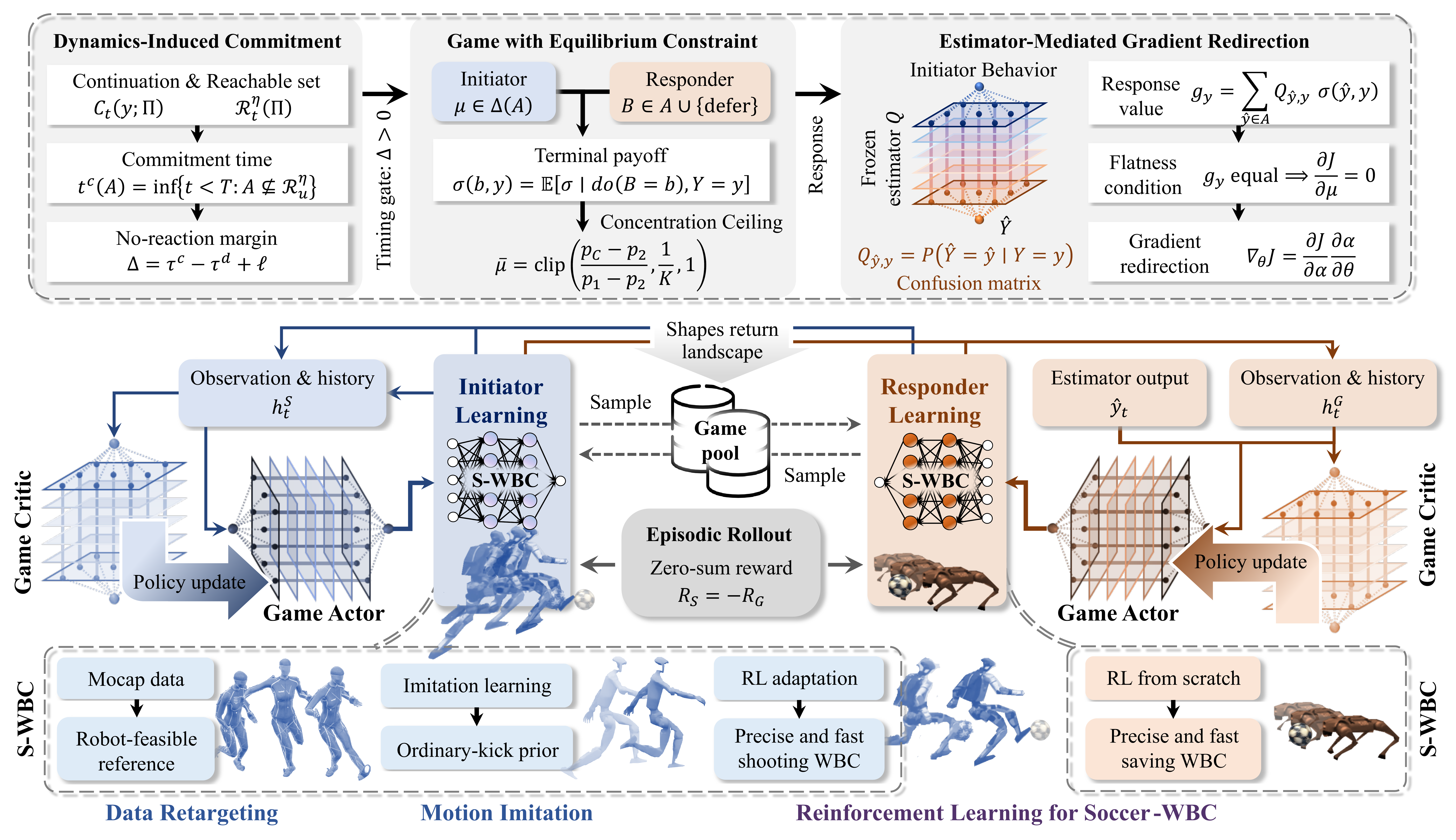}
    \caption{Three-layer architecture separating executable whole-body skills, strategic self-play, and DIC-Map. Fixed S-WBCs realize humanoid shooting and quadruped saving, game-level actor--critics revise strategic commands under zero-sum self-play, and DIC-Map maps the resulting physical rollouts to commitment timing, an equilibrium concentration constraint, and the active first-order learning channel.}
    \label{fig:method_overview}
\end{figure*}

Fig.~\ref{fig:method_overview} separates executable whole-body skills, strategic self-play, and body-grounded game analysis. The fixed S-WBC layer determines what motions remain realizable, and the game layer revises strategic commands. DIC-Map measures when those commands cease to map to distinct terminal outcomes and traces the consequences to equilibrium.

\subsection{Problem Formulation and Hierarchical Architecture}
\label{sec:problem}

The system separates strategic decision making from whole-body execution.
Let $c_t^S$ and $c_t^G$ denote game-level strategic commands, $h_t^S$ and $h_t^G$ the corresponding observation histories, and $u_t^S$ and $u_t^G$ the low-level S-WBC actions.
The responder receives the output $\hat y_t$ of a frozen intent estimator $f_\omega$.
The hierarchy is
\begin{align}
    c_t^S &\sim \pi_{\phi_S}^{\mathrm{game}}(\cdot\mid h_t^S),
    \label{eq:game-layer}\\
    c_t^G &\sim \pi_{\phi_G}^{\mathrm{game}}(\cdot\mid h_t^G,\hat y_t),
    \qquad \hat y_t=f_\omega(h_t^G),\\
    u_t^i &= \pi_{\psi_i}^{\mathrm{S\text{-}WBC}}(o_t^i,c_t^i),
    \qquad i\in\{S,G\},\\
    s_{t+1} &\sim P_{\btheta}(\cdot\mid s_t,u_t^S,u_t^G).
    \label{eq:execution-layer}
\end{align}
where $o_t^i$ is the low-level robot observation, $\phi_i$ parameterizes the game policy, and $\psi_i$ parameterizes the S-WBC.\@
During self-play, the S-WBCs and estimator are fixed and only the game policies are updated.
For opponent policies sampled from the shared pool $\mathcal{P}_{\mathrm{pool}}$, each agent optimizes
\begin{equation}
    J_i(\phi_i)=\E_{\pi_{\phi_i}^{\mathrm{game}},\,\pi_{-i}\sim\mathcal{P}_{\mathrm{pool}}}
    \!\left[\sum_{t=0}^{T-1}\gamma^t R_i\right],
    \qquad R_S=-R_G,
    \label{eq:game-objective}
\end{equation}
using its game-level actor--critic. For the humanoid initiator, self-collected human motion-capture data are retargeted to robot-feasible references. Imitation learning provides an ordinary-kick prior, and reinforcement-learning adaptation produces the shooting S-WBC.\@
For the quadruped responder, the saving S-WBC is learned from scratch by reinforcement learning.
After skill training, $\psi_S$ and $\psi_G$ are frozen in Eqs.~\eqref{eq:game-layer}--\eqref{eq:execution-layer}.
Thus the game policy can revise a strategic command, but only the closed-loop S-WBC trajectory determines whether its terminal alternative remains executable. DIC-Map measures this gap.

Formally, consider a finite-horizon partially observed zero-sum dynamic game $\mathcal{G}_{\btheta}$ over $t\in[0,T]$ between an initiator $S$ and a responder $G$.
The fixed physical parameters $\btheta$ collect morphology, actuation, latency, and environment dynamics, including the closed-loop S-WBC execution.
A state trajectory $s_{0:T}$ induces a terminal outcome $Y=g(s_{0:T})\in\cY$. The responder terminates in a behaviour $B\in A\cup\{\mathrm{defer}\}$ with terminal payoff $\sigma(B,Y)$, while the initiator receives $-\sigma(B,Y)$.
A policy pair induces a terminal allocation $\mu\in\Delta(A)$ with $\mu(y)=\Prob(Y=y)$. The problem is to determine, from this executable hierarchy, when strategic alternatives are physically lost, when the remaining interaction admits a reduced game, and whether policy improvement acts through terminal reallocation or through the responder's estimator.

\subsection{DIC-Map: Commitment and the Reduced Endgame}
\label{sec:commitment}

For a state $s_t$, a class $\Pi$ of continuation controllers for $S$, and a fixed responder probe policy $\pi_G^{\mathrm{probe}}$, define the continuation capability of outcome $y$ by
\begin{equation}
    C_t(y;\Pi)=\sup_{\pi^{\mathrm{cont}}\in\Pi} \Prob\!
    \left(Y=y\mid s_t,\pi^{\mathrm{cont}},
    \pi_G^{\mathrm{probe}},\btheta\right),
    \label{eq:capability}
\end{equation}
and the $\eta$-reachable outcome set by $\cR_t^\eta(\Pi)=\{y\in\cY:C_t(y;\Pi)\ge\eta\}$.
\begin{assumption}[Estimated capability]
    \label{as:capability}
    The true capability $C_t^\star=C_t(\cdot;\Pi_{\mathrm{adm}})$ takes the supremum over all admissible continuations.
    The measured capability $\hat C_t=C_t(\cdot;\hat\Pi)$ uses a stated finite continuation class $\hat\Pi\subseteq\Pi_{\mathrm{adm}}$ under a fixed probe policy.

\end{assumption}

\begin{definition}[Commitment time]
    \label{def:commit}
    Commitment is defined by persistent rather than transient loss. For $A\subseteq\cY$, the commitment time is
    \begin{equation}
        t^c(A)=\inf\!\left\{\,t<T\mid A\not\subseteq\cR_u^\eta\ \text{for all}\ u\in[t,T]\,\right\},
        \label{eq:commitment}
    \end{equation}
    with $t^c=T$ if no such instant exists.
    If $t\mapsto\cR_t^\eta$ is nonincreasing, this reduces to the last time at which all alternatives remain reachable.
\end{definition}

Let $t^d$ be the last time at which the responder can defer without losing its response capability, and let $\ell$ be the initiator's delay in observing and acting on a responder commitment.
Using lead time $\tau=T-t$, define
\begin{equation}
    \Delta=\tau^c-\tau^d+\ell.
    \label{eq:timingmargin}
\end{equation}
\begin{assumption}[Terminal sufficiency, single commitment, and no reaction]
    \label{as:reduction}
    After $t^c$, the payoff depends on the trajectory through $(B,Y)$, and neither player conditions that choice on the opponent's realized terminal choice before its own choice is fixed.
\end{assumption}

Under Assumption~\ref{as:reduction}, the post-commitment interaction is a zero-sum matrix game $\Gamma$ on $\Delta(A\cup\{\mathrm{defer}\})\times\Delta(A)$ with entries $\sigma(b,y)$.

\begin{proposition}[Timing classification]
    \label{prop:role}
    The initiator can react to a realized responder commitment while retaining all alternatives in $A$ if and only if $\tau^d-\ell>\tau^c$, equivalently $\Delta<0$.
\end{proposition}

Thus $\Delta>0$ certifies the initiator side of the no-reaction condition. The responder must still satisfy Assumption~\ref{as:reduction}.

\begin{algorithm}[t]
    \caption{DIC-Map for commitment, equilibrium, and gradient characterization}
    \label{alg:pipeline}
    \begin{algorithmic}[1]
        \Require rollouts, continuation class $\Pi$, probe policy $\pi_G^{\mathrm{probe}}$, threshold $\eta$, latency $\ell$
        \Ensure $\hat t^c$, $\bar\mu$, $\mathrm{Expl}$, active gradient channel
        \State Estimate $\hat C_t(y;\Pi)$ and $\hat{\mathcal R}_t^\eta$
        \State Set $\hat t^c$ by persistent loss of any $y\in A$
        \State Measure $\tau^d$ and compute $\hat\Delta=\hat\tau^c-\tau^d+\ell$
        \If{$\hat\Delta\le0$}
            \State \Return timing not certified
        \EndIf
        \State Estimate $p_1,p_2,p_C$ and compute $r$, $\bar\mu$, and $\mathrm{Expl}$
        \State Estimate $Q$ and $g_y$ from $(\hat Y,Y)$ at responder commitment
        \If{$g_y$ is equal across $y$ within resolution}
            \State \Return estimator-mediated channel
        \Else
            \State \Return allocation + estimator channels
        \EndIf
    \end{algorithmic}
\end{algorithm}

\subsection{Equilibrium Constraint from the Deferral Option}
\label{sec:equilibrium-method}

For $b\in A\cup\{\mathrm{defer}\}$, define the responder's interventional coverage payoff
\begin{equation}
    \sigma(b,y)=\E\!\left[\sigma\mid\doop(B=b),Y=y\right].
    \label{eq:interventional}
\end{equation}
\begin{assumption}[Interventional identification]
    \label{as:interventional}
    An observational conditional $\E[\sigma\mid B=b,Y=y]$ identifies Eq.~\ref{eq:interventional} only when the selected responder behaviour is independent of residual state variation after conditioning on $Y$.
    When this condition is not justified, observational quantities are denoted $\tilde p_1$, $\tilde p_2$, and $\tilde p_C$ and are treated as proxies.
\end{assumption}

\begin{figure}[t]
    \centering
    \includegraphics[width=\columnwidth]{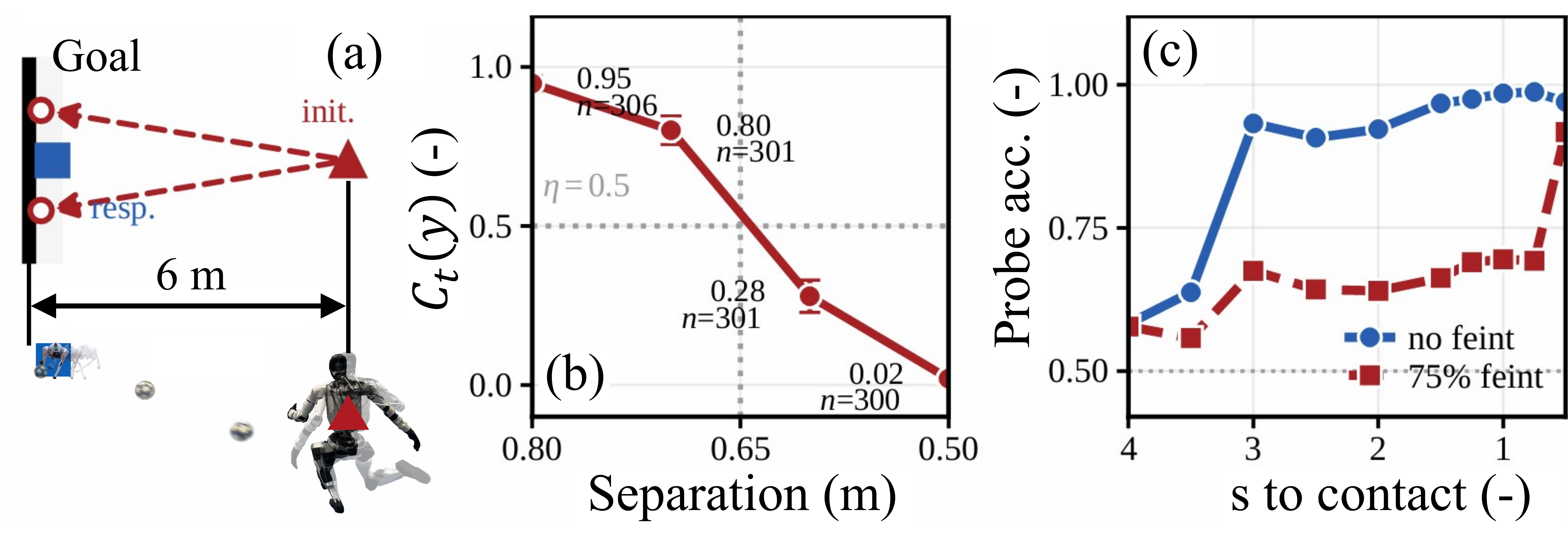}
    \caption{Penalty-game instrumentation showing (a) the terminal aim regions, (b) continuation capability with $95\%$ CIs, where $\eta=0.5$ places $t^c$ at $0.64$\,m or $0.29$\,s before contact, and (c) outcome readability under nominal and scripted revision.}
    \label{fig:instantiation}
\end{figure}

\begin{assumption}[Symmetric coverage]
    \label{as:coverage}
    For every $y\in A$, a matched commitment yields $\sigma(y,y)=p_1$, deferral yields $\sigma(\mathrm{defer},y)=p_C$, and any mismatched commitment yields $\sigma(b,y)=p_2$ for $b\ne y$, with $p_2\le p_C\le p_1$ and $p_2<p_1$.
\end{assumption}
Define the normalized value of deferral by $r=(p_C-p_2)/(p_1-p_2)\in[0,1]$ and the concentration of an initiator allocation by $m(\mu)=\max_{y\in A}\mu(y)$.

\begin{theorem}[Optimal set and concentration ceiling]
    \label{thm:ceiling}
    Under Assumptions~\ref{as:reduction} and~\ref{as:coverage}, the reduced game has value $V^\star=\max\{p_C,p_2+(p_1-p_2)/K\}$, and the initiator's optimal set is $\{\mu:m(\mu)\le\bar\mu\}$ with
    \begin{equation}
        \bar\mu=\operatorname{clip}\!\left(r,\frac{1}{K},1\right),\qquad
        r=\frac{p_C-p_2}{p_1-p_2}.
        \label{eq:ceiling}
    \end{equation}
    A terminal allocation with concentration $m$ concedes
    \begin{equation}
        \mathrm{Expl}(m)=\max\{p_C,p_2+(p_1-p_2)m\}-V^\star.
        \label{eq:expl}
    \end{equation}
\end{theorem}

The ceiling yields three regimes: uniform mixing for $r\le1/K$, partial mixing for $r\in(1/K,1)$, and pure play allowed at $r=1$.
If matched and mismatched committed coverage are fixed while only deferral capability changes, $p_C$ and therefore $\bar\mu$ are nondecreasing in the deferral margin. Fig.~\ref{fig:instantiation} summarizes the penalty-game geometry and the continuation and readability measurements used to instantiate these quantities.

\subsection{Estimator-Mediated Gradient Redirection}
\label{sec:gradient-method}

Suppose the responder does not observe $Y$ directly.
It predicts an outcome $\hat Y\in A$ with confusion matrix $Q_{\hat y,y}=\Prob(\hat Y=\hat y\mid Y=y)$ and, on the estimator-following branch, commits to the predicted alternative.
Define the response value of terminal outcome $y$ as
\begin{equation}
    g_y(Q)=\sum_{\hat y\in A}
    Q_{\hat y,y}\,\sigma(\hat y,y).
    \label{eq:responsevalue}
\end{equation}
The responder's expected payoff from following the estimator is $\sum_y\mu(y)g_y$, whereas deferral yields $p_C$ under symmetric coverage.

\begin{theorem}[Estimator-mediated redirection]
    \label{thm:routing}
    Let $J(\mu,Q)=-\sum_y\mu(y)g_y(Q)$ be the initiator payoff on the estimator-following branch.
    The direct allocation gradient vanishes on the tangent space of $\Delta(A)$ if and only if $g_y$ is identical for all $y\in A$.
    In that case, a policy-induced reallocation has first-order effect
    \begin{equation}
        \frac{\mathrm{d}J}{\mathrm{d}\mu(y)}
        =\underbrace{-g_y}_{\text{equal across outcomes}}
        -\sum_{y'}\mu(y')\frac{\partial g_{y'}}{\partial\mu(y)},
        \label{eq:total}
    \end{equation}
    and only the second term distinguishes reallocation directions.
    Under Assumption~\ref{as:coverage} and a symmetric estimator with $Q_{yy}=\alpha$ and off-diagonal entries $(1-\alpha)/(K-1)$, $g_y\equiv\alpha p_1+(1-\alpha)p_2$.
    Following the estimator is preferable to deferral exactly when $\alpha>r$.
    If $\alpha\ge1/K$, the effective threshold is
    \begin{equation}
        \begin{gathered}
            \alpha^\star_{\mathrm{eff}}
            =\operatorname{clip}(r,\tfrac1K,1)=\bar\mu,\\
            \frac{\partial J}{\partial\mu}=0,\qquad
            \frac{\partial J}{\partial\alpha}=-(p_1-p_2)<0.
        \end{gathered}
        \label{eq:gradients}
    \end{equation}
\end{theorem}

\begin{figure*}[t]
    \centering
    \includegraphics[width=\textwidth]{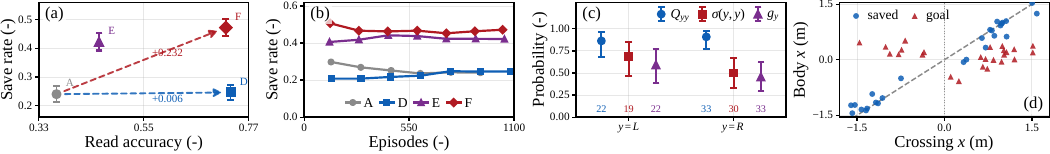}
    \caption{Controlled responder comparison at $\beta=0.8$ showing (a) read accuracy versus save rate, where A$\rightarrow$D and A$\rightarrow$F yield similar accuracy gains but save-rate gains of $+0.006$ and $+0.232$, respectively, (b) cumulative save rate, (c) policy-F class-conditional $Q_{yy}$, $\sigma(y,y)$, and $g_y$, and (d) terminal responder position versus ball crossing.}
    \label{fig:branch}
\end{figure*}

\begin{figure}[t]
    \centering
    \includegraphics[width=\columnwidth]{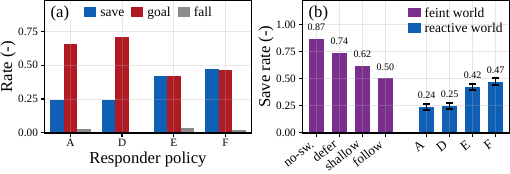}
    \caption{Responder outcomes across evaluation settings showing (a) save, goal, and fall rates for policies A, D, E, and F and (b) a scripted-feint comparison in which deferral attains a save rate of $0.736$ versus $0.502$ for early-read following, with the reactive-world bars reported using $95\%$ CIs.}
    \label{fig:crossplay}
\end{figure}

\begin{table}[t]
    \centering
    \caption{Responder ablation at $\beta=0.8$ against the same reactive initiator, reporting episode count, read accuracy, and save rate with $95\%$ CIs.}
    \label{tab:responder-comparison}
    \tablefont
    \renewcommand{\arraystretch}{1.08}
    \setlength{\tabcolsep}{2.7pt}
    \begin{tabular}{@{}l l c c c@{}}
        \toprule
        Policy & Variant & Ep. & Read & Save rate $[95\%\ \mathrm{CI}]$ \\
        \midrule
        A & Base & $924$ & $0.367$ & $0.240\ [0.212,0.268]$ \\
        D & Deferral & $1081$ & $0.732$ & $0.246\ [0.220,0.272]$ \\
        E & Deferral+decoy & $1050$ & $0.456$ & $0.422\ [0.392,0.452]$ \\
        F & DAgger estimator & $1041$ & $0.722$ & $0.472\ [0.442,0.502]$ \\
        \bottomrule
    \end{tabular}
\end{table}

\begin{figure}[t]
    \centering
    \includegraphics[width=\columnwidth,trim={10.5 0 11 11},clip]{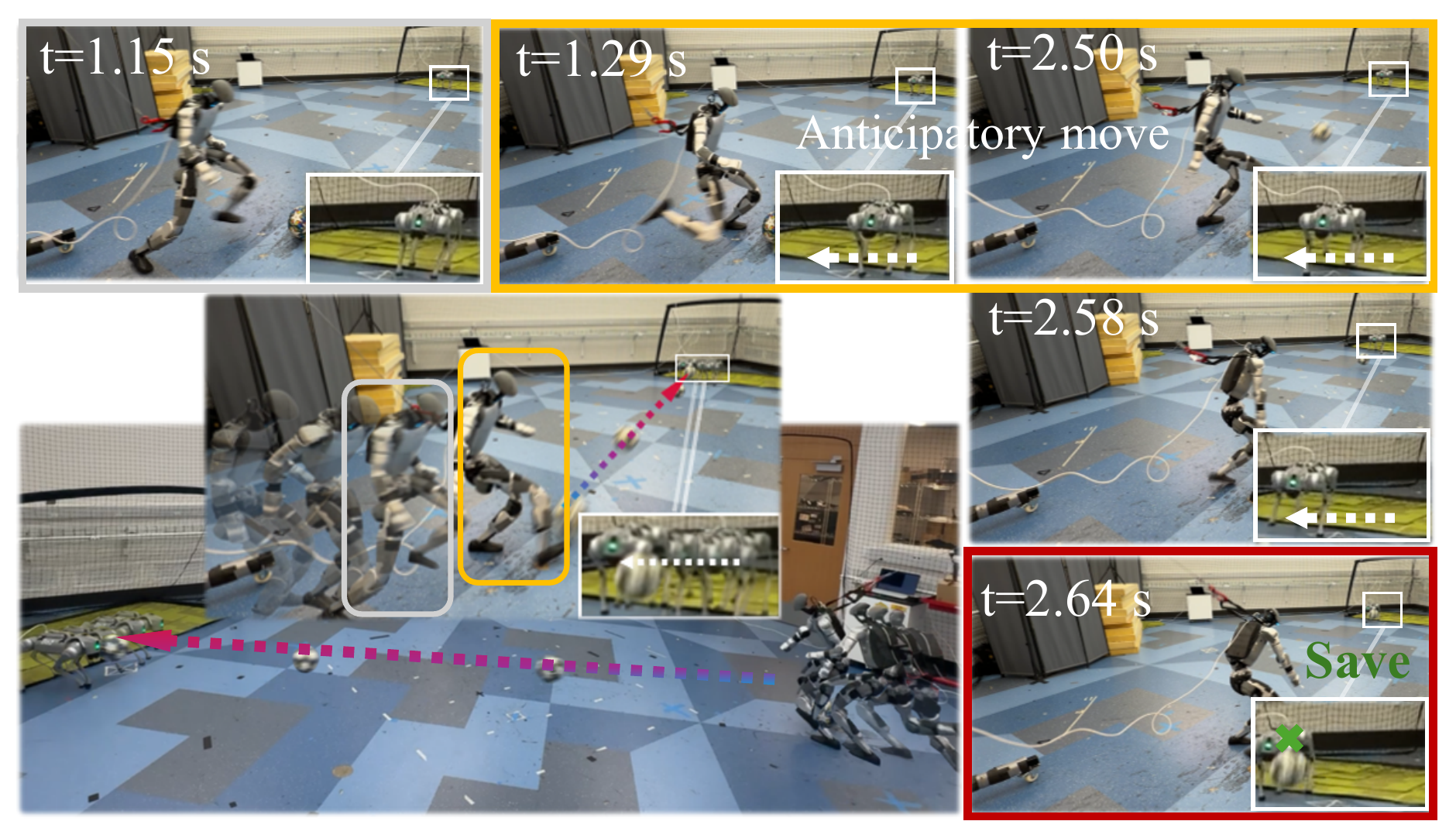}
    \caption{Hardware deployment of anticipatory goalkeeping in which Go2 predicts G1's final shot side during kick preparation, initiates an early interception motion, and blocks the ball.}
    \label{fig:qualitative-left-save}
\end{figure}

In the interior regime $r\in(1/K,1)$, the identity $\alpha^\star_{\mathrm{eff}}=\bar\mu$ links the estimator-accuracy threshold directly to the equilibrium concentration ceiling. The frozen estimator's effective confusion matrix changes only because the initiator changes the trajectory distribution.

\begin{corollary}[Redirection in policy-parameter space]
    \label{cor:routing}
    Let $\phi_S$ parameterize the initiator game policy, with induced terminal allocation $\mu(\phi_S)$ and symmetric-estimator accuracy $\alpha(\phi_S)$.
    On the estimator-following branch,
    \begin{equation}
        \nabla_{\phi_S} J
        =\underbrace{\frac{\partial J}{\partial\mu}
            \frac{\partial\mu}{\partial\phi_S}}_{=0}
        +\frac{\partial J}{\partial\alpha}
        \frac{\partial\alpha}{\partial\phi_S}
        =\frac{\partial J}{\partial\alpha}
        \frac{\partial\alpha}{\partial\phi_S}.
        \label{eq:chain}
    \end{equation}
\end{corollary}

The estimator-mediated mechanism is evaluated by the controlled responder ablation in Fig.~\ref{fig:branch}, the cross-world comparison in Fig.~\ref{fig:crossplay}, and the common-protocol statistics in Table~\ref{tab:responder-comparison}. Fig.~\ref{fig:qualitative-left-save} shows a hardware interception. Algorithm~\ref{alg:pipeline} summarizes DIC-Map.

\section{Results and Discussion}
\label{sec:results}

\subsection{Experimental Setup and Measurement Protocol}
\label{sec:setup}

We instantiate the hierarchical system in Isaac Sim/Isaac Lab \cite{mittal2023orbit} with parallel simulation \cite{rudin2022legged}.
The $29$-degree-of-freedom humanoid initiator strikes toward a goal plane $7.5$\,m away, and the quadruped responder attempts to save.
Two aim regions centered at $\pm0.85$\,m define the terminal alternatives $A$.
A scalar $\beta$ scales launch velocity, changing responder time.
The S-WBCs and responder estimator are fixed throughout the strategic comparisons.

The responder reward is $+5$ for a stop and $-5$ for a concession, with no shaping on prediction, switching, deception, commitment, or style. Thus switching and feint-like interactions are not directly rewarded.
We analyze v1 at $\beta=0.8$ and v2 at $\beta=0.9$.
Continuation capability is estimated by branching from stored physical states, commanding alternative terminal targets, and recording realized outcomes. The primary threshold is $\eta=0.5$. All strategic quantities are scored from realized physical state rather than the $10$\,Hz decision variable because commands can change without corresponding body motion. The same body-grounded convention is used for terminal outcomes, responder occupancy, commitment ordering, and coverage.

Three protocols test the mechanisms below.
The continuation probe estimates $\hat C_t$ and $\hat t^c$.
A fixed-policy replay changes only $\beta$ to test deferral coverage.
A reactive-initiator protocol at $\beta=0.8$ compares four responder policies in $32$ parallel environments for $360$\,s: A is the base commit-on-read policy, D adds deferral, E adds a decoy, and F replaces only the estimator.
Policy F records $(\hat Y,Y)$ and terminal body state for class-conditional analysis.

\begin{figure*}[t]
    \centering
    \includegraphics[width=\textwidth]{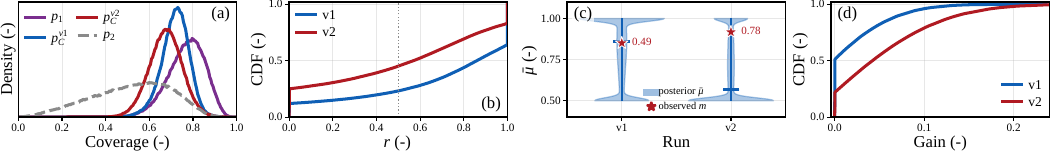}
    \caption{Posterior equilibrium analysis showing (a) coverage posteriors, (b) the posterior CDF of $r$, (c) the concentration-ceiling posterior against observed $m$, and (d) the posterior CDF of deviation gain. The broader downstream uncertainty reflects limited support for the mismatched-commitment cell.}
    \label{fig:ceiling}
\end{figure*}

\begin{table*}[t]
    \centering
    \caption{Posterior concentration-ceiling comparison across league runs obtained by propagating Jeffreys coverage posteriors through Eqs.~\ref{eq:ceiling}--\ref{eq:expl}.}
    \label{tab:equilibrium-posterior}
    \tablefont
    \renewcommand{\arraystretch}{1.08}
    \setlength{\tabcolsep}{7.0pt}
    \begin{tabular*}{\textwidth}{@{\extracolsep{\fill}}l cc ccc@{}}
        \toprule
        & \multicolumn{2}{c}{Observed} & \multicolumn{3}{c}{Posterior prediction} \\
        \cmidrule(lr){2-3}\cmidrule(lr){4-6}
        Run & $\tilde p_C$ & $m$ & $\bar\mu\ [95\%\ \mathrm{CrI}]$ & Gain $[95\%\ \mathrm{CrI}]$ & $\Prob(m>\bar\mu)$ \\
        \midrule
        v1 & $0.741$ & $0.850$ & $0.86\ [0.50,1.00]$ & $0.000\ [0.000,0.112]$ & $0.51$ \\
        v2 & $0.667$ & $0.917$ & $0.57\ [0.50,1.00]$ & $0.042\ [0.000,0.191]$ & $0.78$ \\
        \bottomrule
    \end{tabular*}
\end{table*}

\begin{figure}[t]
    \centering
    \includegraphics[width=\columnwidth,trim={1 5.5 6.5 1},clip]{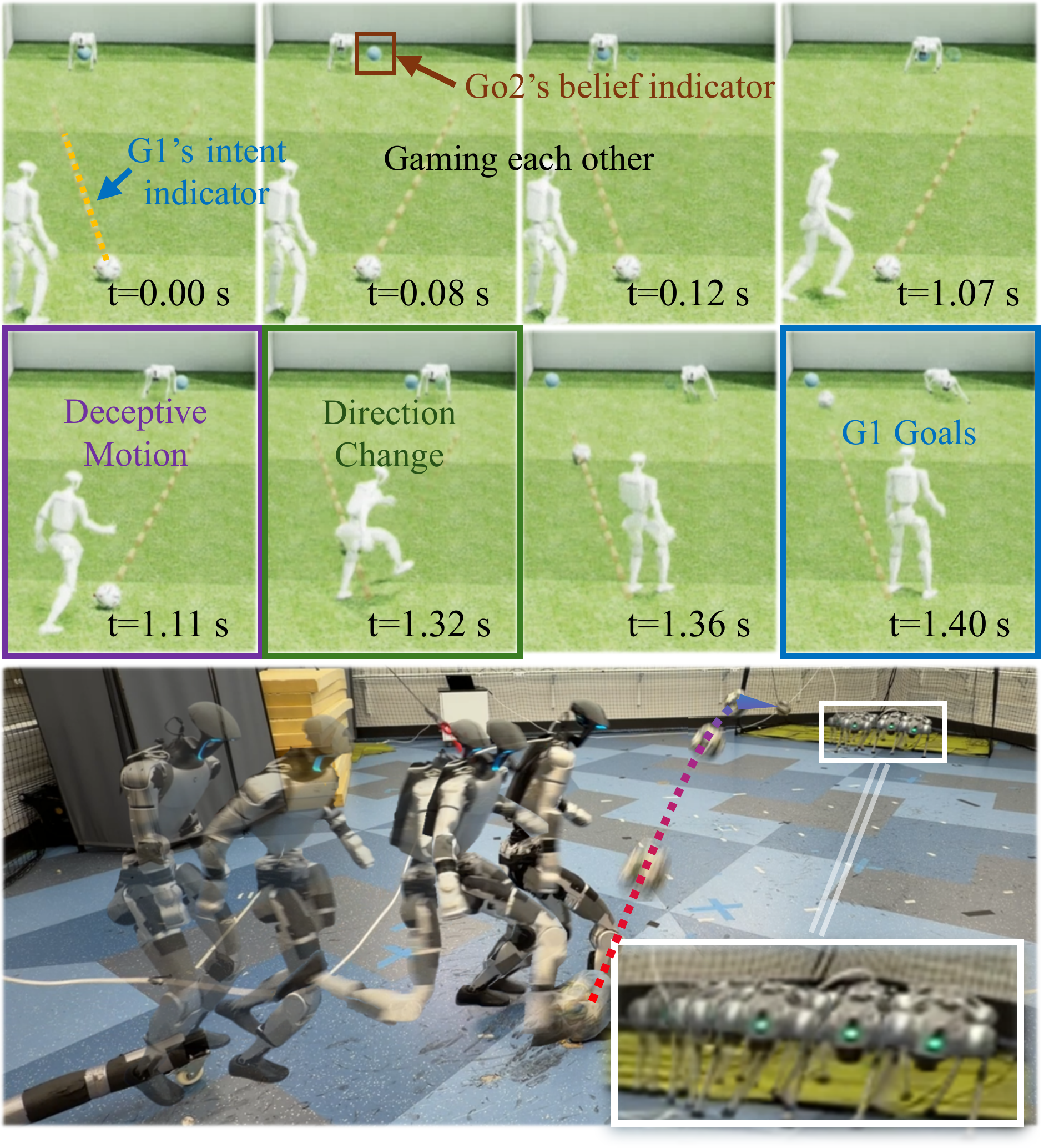}
    \caption{Continuous strategic interaction in which G1 changes its shot direction immediately before ball contact, causing Go2 to move toward the wrong interception location and concede, with the upper sequence showing simulation and the lower sequence the corresponding hardware trial.}
    \label{fig:qualitative-feint}
\end{figure}

\begin{figure}[t]
    \centering
    \includegraphics[width=\columnwidth]{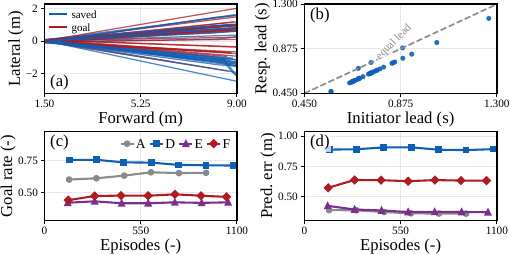}
    \caption{Policy-F state recording showing (a) ball trajectories grouped by terminal event, (b) responder read-lock time against the initiator commitment time, where every sampled episode locks after the ball side is fixed, (c) goal rate, and (d) lateral prediction error.}
    \label{fig:staterec}
\end{figure}

\begin{figure}[t]
    \centering
    \includegraphics[width=\columnwidth]{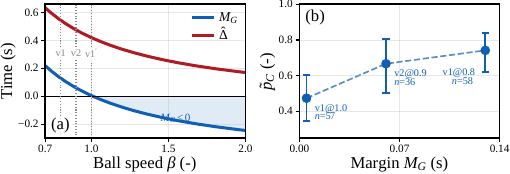}
    \caption{Physical timing across ball speed showing (a) the responder spare-time margin $\MG$ and no-reaction margin $\hat\Delta$ and (b) measured deferral coverage versus $\MG$ with $95\%$ Jeffreys intervals and cell counts.}
    \label{fig:phase}
\end{figure}

The responder spare-time margin is $\MG=t_{\mathrm{flight}}-t_G^{\mathrm{cover}}$, whereas $\Delta$ in Eq.~\ref{eq:timingmargin} tests whether the initiator can react after responder commitment.
Binomial rates use $95\%$ confidence intervals. Jeffreys posteriors are propagated through Eqs.~\ref{eq:ceiling}--\ref{eq:expl} for sparse coverage cells.
Observational coverage proxies are marked with a tilde.
Table~\ref{tab:equilibrium-posterior} reports the run-level posterior quantities used in the equilibrium comparison. Fig.~\ref{fig:qualitative-feint} shows a late-revision interaction, Fig.~\ref{fig:staterec} records policy-F trajectories and commitment ordering, and Fig.~\ref{fig:phase} relates ball speed to the physical timing margins and deferral coverage.

\subsection{Estimator-Mediated Response Improvement}
\label{sec:res-channel}

The controlled responder comparison separates information quality from response value by asking whether a comparable gain in read accuracy is equally useful when obtained through additional waiting or through a better estimator.
Moving from A to D raises read accuracy from $0.367$ to $0.732$ but changes save rate only from $0.240$ to $0.246$.
Moving from A to F produces a similar accuracy increase, to $0.722$, while save rate rises to $0.472$.
Thus the two changes improve raw read accuracy by $0.365$ and $0.355$, respectively, but change save rate by only $+0.006$ through deferral and by $+0.232$ through the estimator replacement.

\begin{figure}[t]
    \centering
    \includegraphics[width=\columnwidth]{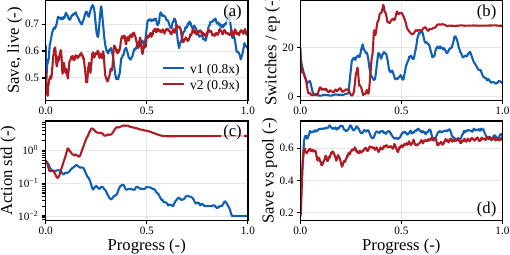}
    \caption{League training dynamics for the two analyzed runs showing (a) live save rate, (b) initiator switches per episode, (c) responder action standard deviation, and (d) save rate against the frozen archive, which increases from $0.51$ to $0.68$.}
    \label{fig:training}
\end{figure}

\begin{figure}[t]
    \centering
    \includegraphics[width=\columnwidth,trim={15 1.5 1.5 1.5},clip]{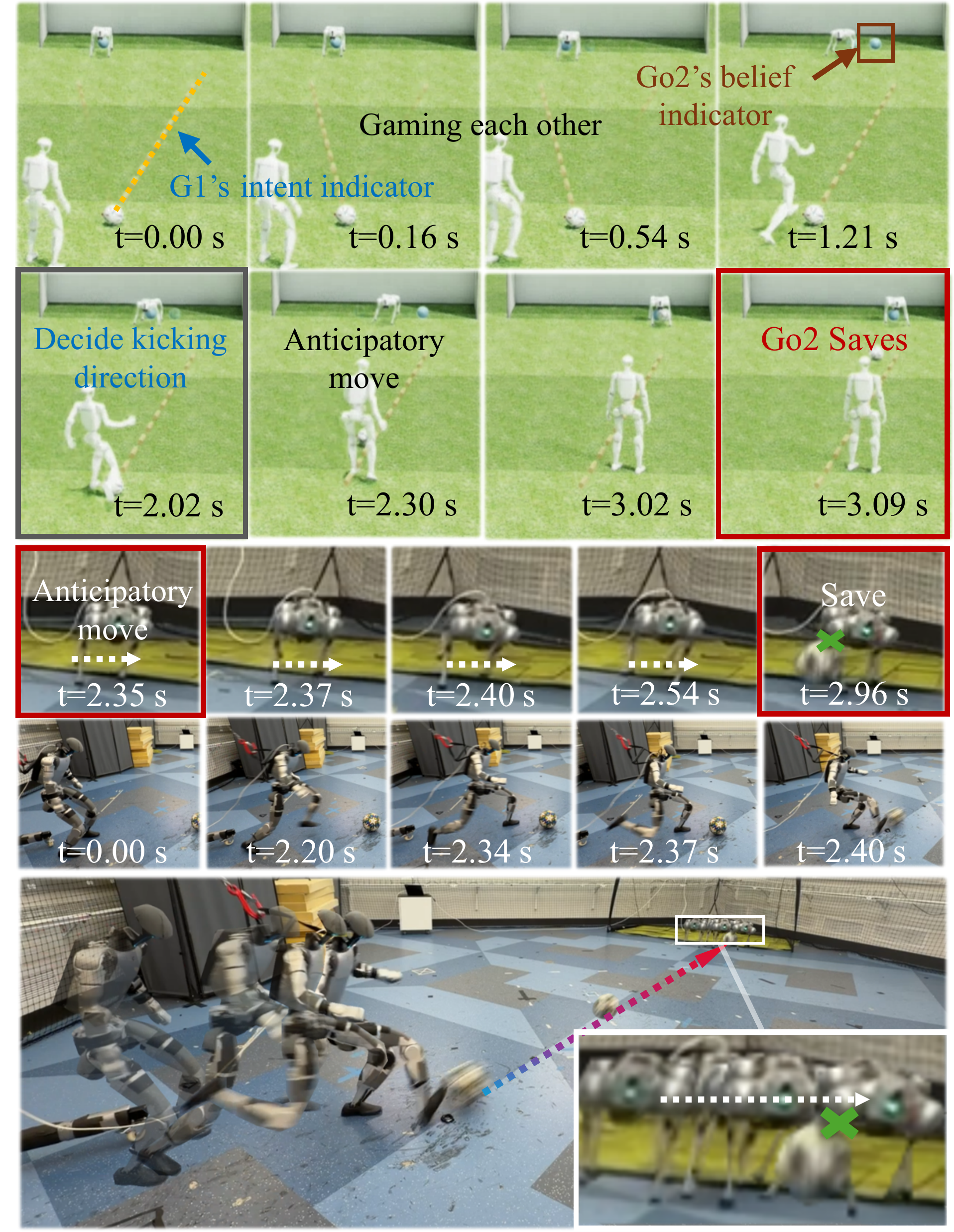}
    \caption{Successful anticipatory goalkeeping during continuous strategic interaction, where Go2 predicts G1's final shot direction despite late target revision and saves the ball, with the top row showing simulation, the middle row showing separately selected and vertically unsynchronized hardware key frames for each robot, and the bottom row showing the hardware overview.}
    \label{fig:qualitative-right-save}
\end{figure}

The scripted-feint comparison further shows the value of deferral, with a save rate of $0.736$ versus $0.502$ for following the early read. We use this test as a cross-world response comparison rather than an equilibrium metric. The two class-conditional diagonal accuracies are $0.864$ and $0.909$.
Combining the confusion matrix with the measured side-conditioned response gives $g_L=0.591$ and $g_R=0.455$. Their difference is not statistically resolved (Fisher $p=0.41$), so we treat them as compatible with response-value flatness at this resolution.
Together with Fig.~\ref{fig:branch} and Table~\ref{tab:responder-comparison}, these results show that read accuracy alone does not determine save performance.

\subsection{Dynamics-Induced Commitment and Deferral Capability}
\label{sec:res-primitive}

The continuation probe estimates when terminal alternatives are lost.
Measured continuation capability for a commanded target revision decreases from $0.948$ at a foot--ball separation of $0.8$\,m to $0.020$ at $0.5$\,m, with most of the transition between $0.7$ and $0.6$\,m.
At $\eta=0.5$, the crossing occurs at approximately $0.64$\,m, corresponding to a commitment lead of $0.29$\,s.
The estimated lead changes by $35$\,ms as $\eta$ varies from $0.3$ to $0.7$.
Readability and commitment are separated in time. The terminal side is read at approximately $0.92$ accuracy half a second before contact, while revision remains possible until roughly $0.29$\,s. This readable-but-reversible interval permits strategic anticipation and late revision.
Under Assumption~\ref{as:capability}, the measured commitment lead is an upper bound on the true lead because the continuation class is finite.

The fixed-policy replay tests how dynamics affect the responder's outside option.
Increasing $\beta$ from $0.8$ to $1.0$ reduces the responder spare-time margin from $0.130$ to $0.005$\,s and lowers measured deferral coverage from $0.741$ ($n=58$) to $0.474$ ($n=57$).
The policies and terminal regions are unchanged, so this manipulation varies the physical time available to defer while holding the strategic mapping fixed.
This matches the reduced-game monotonicity under fixed committed coverage, where a smaller deferral margin lowers the outside option entering the concentration ceiling.

The no-reaction margin remains positive over the evaluated ball speeds: $\hat\Delta=0.542$\,s, $0.473$\,s, and $0.417$\,s for $\beta=0.8$, $0.9$, and $1.0$.
These values support the initiator side of the reduced-game timing condition for the evaluated regime. Because $\hat\tau^c$ is an estimated upper bound, this timing result applies to the evaluated continuation class rather than the full admissible class.
On realized Policy-F trajectories, the ball side becomes fixed at a lead of $0.751\pm0.126$\,s, while the responder's read lock occurs at $0.659\pm0.130$\,s.

\subsection{Equilibrium Constraint and Learned Play}
\label{sec:res-equilibrium}

The two league runs connect the physically measured deferral value to the strategic allocation learned under self-play.
On physical-state traces, the dominant terminal alternative has probability $0.850$ in v1 and $0.917$ in v2.
Across the same runs, commanded within-episode target revisions rise from $4.74$ to $28.94$ per episode over $1631$ and $1949$ evaluation episodes. The measured coverage cells are $\tilde p_1=0.762$ ($n=21$), $\tilde p_C^{\mathrm{v1}}=0.741$ ($n=58$), $\tilde p_C^{\mathrm{v2}}=0.667$ ($n=36$), and $\tilde p_2=0.667$ ($n=3$).
Because these values are conditioned on the responder's selected behaviour, we treat them as observational proxies rather than identified interventions. We therefore use the propagated posterior as a directional equilibrium test.
Relative to v1, v2 has lower deferral coverage with posterior probability $0.78$, a lower ceiling with probability $0.87$, and larger deviation gain with probability $0.74$. For v2, $\Prob(m>\bar\mu)=0.78$.
The hardware sequence in Fig.~\ref{fig:qualitative-right-save} provides an instance of continuing strategic interaction at deployment. G1 revises intended direction late in the sequence, while Go2 still anticipates the final side and saves the shot.

\section{Conclusions}
\label{sec:conclusion}

We presented a body-grounded DIC-Map framework showing that, in hierarchical robotic games, whole-body dynamics can change the strategic game before either agent explicitly commits. The results show that whole-body dynamics is not just execution constraint. They determine which strategies remain available, which reduced game is induced, and which first-order learning channel remains effective. The main contributions are summarized as follows:
\begin{itemize}
    \item DIC-Map maps executable S-WBC rollouts to continuation capability, persistent commitment time, and a reduced endgame tested with timing. In the evaluated task, the shot side becomes highly predictable before the humanoid loses the physical ability to revise it, with commitment about $0.29$\,s before contact.
    \item The reduced symmetric game yields a closed-form concentration ceiling governed by the responder's value of deferring. Changing only ball speed changes deferral coverage while the strategic action space and policies are held fixed, supporting the predicted direction of the physical-to-equilibrium link.
    \item Under equal response values, the direct terminal-allocation gradient vanishes and the estimator-mediated channel remains. Replacing the estimator raises save rate from $0.240$ to $0.472$, whereas a comparable gain in read accuracy obtained by waiting reaches only $0.246$. The learned interaction exhibits frequent within-episode revision without behaviour-specific shaping, and hardware trials show both a late revision that misdirects the responder and a successful anticipatory save.
\end{itemize}

\FloatBarrier
\bibliographystyle{IEEEtran}
\bibliography{references}

\end{document}